# Retrieval and Generative Approaches for a Pregnancy Chatbot in Nepali with Stemmed and Non-Stemmed Data : A Comparative Study

Sujan Poudel*, Nabin Ghimire†, Bipesh Subedi‡ and Saugat Singh§
Department of Computer Science and Engineering, Kathmandu University
Dhulikhel, Nepal
Email: *poudelsujan03@gmail.com, †nabin.ghimire@ku.edu.np, ‡bipeshrajsubedi@gmail.com, §saugatsingh09@gmail.com

***Abstract*—The field of Natural Language Processing which involves the use of artificial intelligence to support human languages has seen tremendous growth due to its high-quality features. Its applications such as language translation, chatbots, virtual assistants, search autocomplete, and autocorrect are widely used in various domains including healthcare, advertising, customer service, and target advertising. To provide pregnancy-related information a health domain chatbot has been proposed and this work explores two different NLP-based approaches for developing the chatbot. The first approach is a multiclass classification-based retrieval approach using BERT-based multilingual BERT and multilingual DistilBERT while the other approach employs a transformer-based generative chatbot for pregnancy-related information. The performance of both stemmed and non-stemmed datasets in Nepali language has been analyzed for each approach. The experimented results indicate that BERT-based pre-trained models perform well on non-stemmed data whereas scratch transformer models have better performance on stemmed data. Among the models tested the DistilBERT model achieved the highest training and validation accuracy and testing accuracy of 0.9165 on the retrieval-based model architecture implementation on the non-stemmed dataset. Similarly, in the generative approach architecture implementation with transformer 1 gram BLEU and 2 gram BLEU scores of 0.3570 and 0.1413 respectively were achieved.**

## I. Introduction

Technology's influence continues to reshape various sectors that simplify lives and drive progress. This evolution is especially evident in the healthcare domain, which is experiencing transformative changes due to advancements in technology. Among these advancements, healthcare assistants have emerged as versatile tools. One notable innovation within this field is the integration of chatbot technology. his research focuses on using chatbots to address pregnancy-related inquiries, especially in regions with limited healthcare access. By developing a Nepali language-based chatbot specialized in providing pregnancy information, we aim to bridge the healthcare knowledge gap and improve accessibility for individuals seeking guidance during this critical period of their lives.

Chatbots is a computer programs designed to simulate human-like conversations. This potential is particularly promising in offering valuable pregnancy-related information to individuals. The term "chatbot" itself, coined by merging "chat" and "robot," underscores their role as conversational agents powered by technology.

The roots of chatbot technology trace back to the development of ELIZA [1]. These chatbots, whether focused on question-answering or open-ended conversation, are gaining traction across diverse domains like healthcare, e-commerce, finance, and hospitality. This surge is attributed to their effectiveness in engaging users in human-like interactions.

Two primary categories of chatbots exist: rule-based and NLP-based. Rule-based chatbots rely on predefined patterns and decision trees, while NLP-based chatbots leverage Artificial Intelligence (AI) techniques to understand and respond to human language. The rise of Natural Language Processing (NLP) opens doors to AI-powered chatbots, which are gaining attention for their potential in delivering accurate and domain-specific responses. This open domain language model has garnered attention for its exceptional performance, showcasing the possibilities for highly accurate and task-specific chatbots.

This study introduces a pregnancy-related question-answer chatbot designed for the Nepali language. By scraping data from medical websites FAQs, the aim of this study is to provide reliable information to users. These experiments employ AI-powered approaches, including retrieval and generative techniques, utilizing models like BERT and transformer architecture. The performance of these approaches is evaluated on stemmed and non-stemmed datasets that highlight trends in effectiveness.

In regions like Nepal, where healthcare access, particularly during pregnancy, is a challenge, technology like chatbots can play a pivotal role. By providing information through familiar platforms like smartphones, we can contribute to bridging the healthcare gap. This research strives to lay the foundation for a domain-specific chatbot that caters to Nepali speakers, addressing their unique needs and providing crucial healthcare support.

## II. Related Works

Several medical chatbot models have been introduced in the literature, each with distinct features and purposes. One such model, Diabot [2], utilized ensemble learning techniques to enhance its predictive capabilities in health-related scenarios . Employing the RASA NLU platform, Diabot tackled both general health prediction and advanced diabetes prediction tasks. The dataset encompassed a generic health dataset for the former and the Pima Indian diabetes dataset for the latter. Notably, stacking-based ensemble learning with majority voting was employed in training. An approach on development of chatbot to encourage healthy living [3], trained using SVM, Naive Bayes, and KNN algorithms, the chatbot provided advice and remedies based on symptoms. Similarly, a pattern-based chatbot for diagnosis [4], remedy suggestions, and doctor referrals. Both of these models utilized AIML format datasets. Jollity [5], a generative-based chatbot for supporting individuals dealing with stress. Built using the Rasa framework, the chatbot corrected spelling errors and provided motivational content.

A medical chatbot [6] employing artificial intelligence processed user questions and answers using techniques like tokenization, stop word removal, n-gram, TF-IDF, and cosine similarity ranking.A multilingual healthcare chatbot [7] was developed by researchers for early disease diagnosis, using techniques like TF-IDF and Cosine Similarity. MedBot [8], focused on COVID-19 assistance and advice. Using DialogFlow, this chatbot offered medical guidance and referred patients to doctors when necessary. A COVID-19 assistance chatbot, Health Bot [9], was designed to offer resources and connections during the pandemic . It employed Python, machine learning, and deep learning concepts, operating as a retrieval-based chatbot. A COVID-19 informative chatbot [10] based on the BERT language model for question answering. Masum et al. [11] used a transformer model to create a Bengali general knowledge chatbot, achieving high BLEU scores. Similar development has been observed on a web-based college information chatbot [12] with student interaction in mind. While Pearce have provided an open-source tool for students and teachers to create transformer-based chatbots using course materials, allowing customization and learning about AI principles. This tool guided users through various stages of model creation. Each of these approaches contributes to the expanding landscape of medical chatbots, catering to a wide array of healthcare needs [13].

## III. Methodology

The procedures that have been followed to accomplish this research work involves data collection, dataset preparation followed by model design, implementation along with performance analysis. Figure 1 depicts the procedure followed to perform this study.

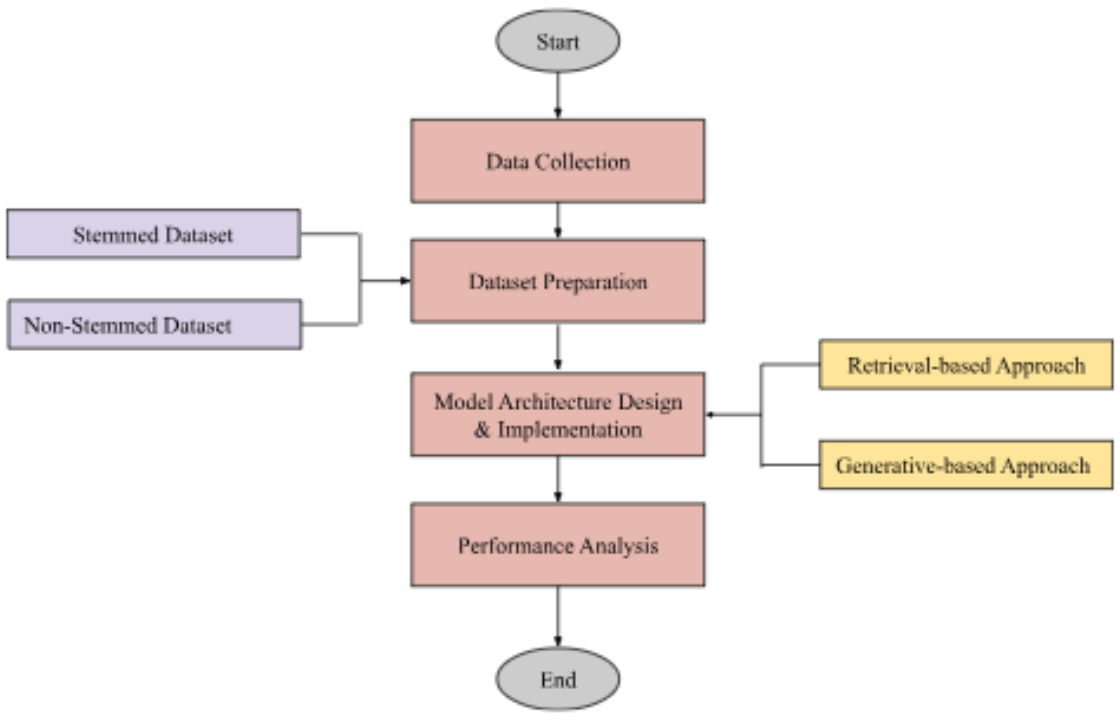

Figure 1. Research Design Flow.

### A. Data Collection

Chatbots become more intelligent as they have access to more knowledge. The data collection process is called dataset collection. There are various methods that can be used to gather dataset for question answer bot or chatbot are scrapping FAQ from the website, user interactions ( questionnaire, survey etc), use existing dataset and improve, collect data from external sources such as databases, APIs, and other online resources and so on. The corpora of data is retrieved by using an information retrieval technique called scraping by using a python library called BeautifulSoup from different government websites which have information and FAQs related to pregnancy. The websites that have been scraped for pregnancy related data are websites of Public Health England, World Health Organization (WHO), National Institute of Child Health and Human Development, Ministry of Health and Wellness Bahamas and so on. Overall, data collection is an important aspect of chatbot development, as it allows chatbots to learn and improve over time. The data collection through the above mentioned method leads to collection of 1159 different questions and its answers in the English language. Among them some questions were manually added to the dataset in each category by taking analogies in the FAQ related to pregnancy.

### B. Dataset Preparation

The data preparation phase is a critical stage in constructing a suitable dataset. The process initiated by eliminating duplicate entries from the repository of pregnancy-related queries and responses, resulting in 817 unique pairs. To make the dataset bigger and stronger, different methods were used to create more examples, giving a total of 21,519 pairs in 817 categories. These pairs were then translated into Nepali with googletrans API. The dataset was also cleaned up by removing extra spaces, empty spots, and punctuation marks, and mixing of English and Nepali characters. The Porter stemming algorithm have been used to perform stemming operation or to figure out root words in the questions and took out words that didn't matter. After this, unnecessary questions were removed that leaves 18,925 questions in 817 groups. Lastly, the dataset was divided into parts for training, validation, and final testing for 70% , 20% and 10% respectively. The dataset before stemming and after stemming are shown in figure ?? and figure ?? below:

### C. Data Preprocessing

The data preparation was followed by formatting it in a way that is compatible with the transformer model. Tokenization was identified as a key step in this process, as it enabled the conversion of the text into a sequence of tokens that can be input into the transformer model. To ensure that the transformer model can process the input sequences a maximum length of 250 words was decided to make all input sequences uniform in length. The maximum input sequence length has been decided after considering the length of the longest sentences in the dataset was found 192 so that a maximum length of 250 words would be sufficient to cover the majority of the sentences while also keeping the computational cost reasonable. Furthermore, a vocabulary was created and the tokenized data was converted into numerical values for processing during model training.

### D. Model Architecture Design and Implementation

In this section, the architecture used for building the question answer bot related to pregnancy has been discussed. Two different approaches, namely retrieval-based approach and generative-based approach, have been implemented in this research. The explanation of these approaches is given below.

#### 1) Model Design:

*a) Retrirval based approach:* The performance of the rule-based chatbot was found to be inferior to that of the NLP-based chatbot. Hence, a multiclass classification approach was implemented for the retrieval-based method by leveraging NLP concepts. To achieve this, state-of-the-art encoder-based transformer pre-trained models such as multilingual BERT [14] and multilingual DistilBERT [15] were utilized, which have been demonstrated to improve the chatbot's performance. The architecture of the question answer bot related to pregnancy using retrieval based approach is depicted in figure 2

*b) Generative based approach:* The generative approach for chatbot development utilizes deep learning techniques to produce human-like conversations or text. One of the most well-known open-domain chatbots called chatGPT utilizes the transformer architecture known as Generative Pre-trained Transformer (GPT-3). Various techniques have been proposed in literature for building generative chatbots. However, the transformer based encoder-decoder architecture has been implemented for the pregnancy-related question-answer bot as it is currently the most advanced deep learning architecture in the natural language processing domain. The figure 3 depicts the architecture of pregnancy related QA bot with generative approach.

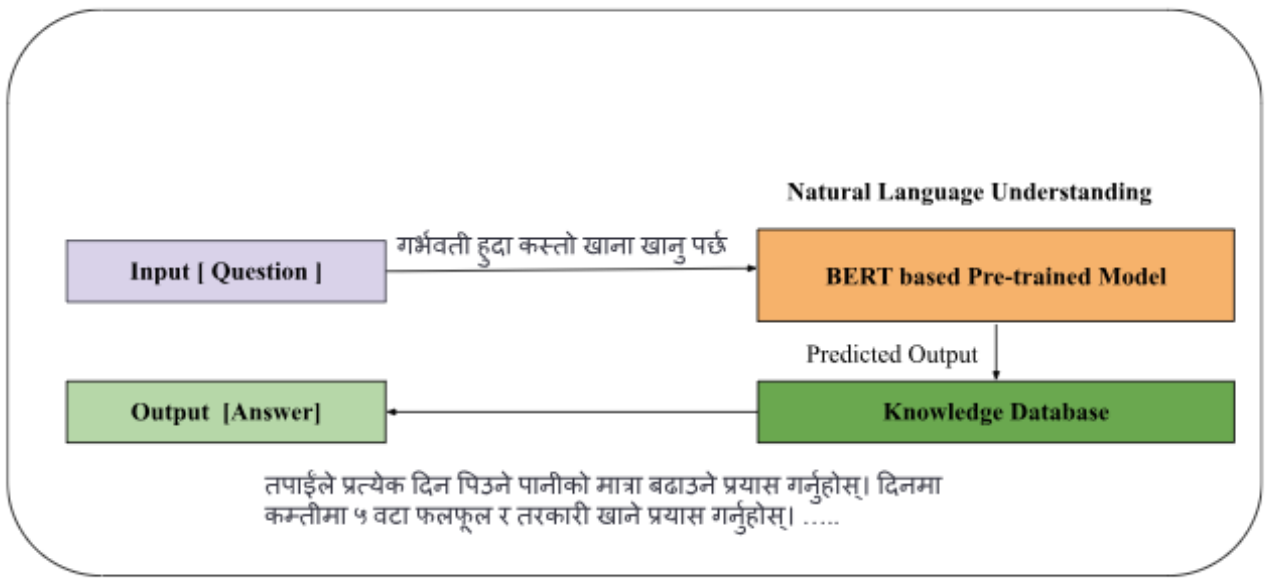

Figure 2. Pregnancy related QA bot architecture for retrieval based approach

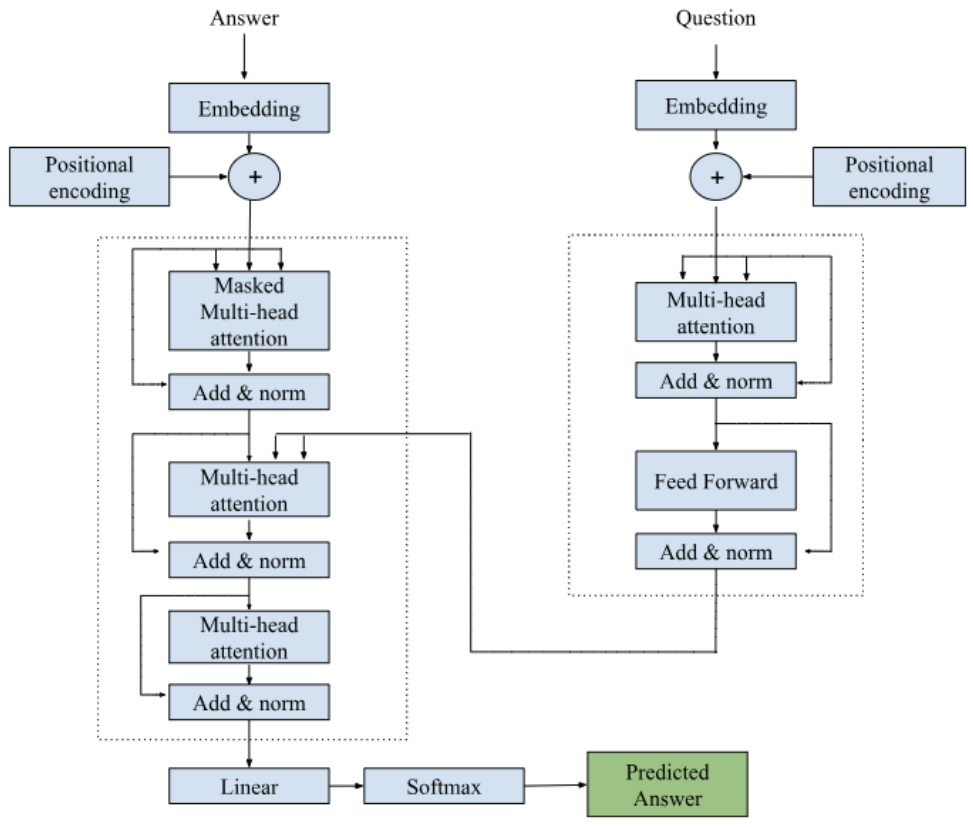

Figure 3. Generative approach pregnancy related QA bot architecture

#### 2) Model Implementation:

*a) Model Training:* In Model A implementation, the pre-trained multilingual BERT or DistilBERT model is fine-tuned using the provided 'question' and its corresponding categories. The model is trained to learn the patterns and associations between the questions and their respective categories in the training data. This process involves feeding the preprocessed 'question' and its corresponding category to the model as input.

Similarly, Model B architecture was trained using preprocessed data with question and answer parameters, after distributing the dataset into training, validation, and testing sets. During the training process, the preprocessed questions were passed as input vector sequences through the encoder, while the decoder took the corresponding answer vector sequences. The transformer's positional encoding feature ensured that the input vector's position was taken into account. The transformer's self-attention mechanism calculated and identified the importance of different parts of the input using a scaled dot product.

Table I
Proposed model architectures Implementation

| Model | Architecture |
|---|---|
| Model A | Retrieval Approach [ Multilingual BERT/ DistilBERT ] |
| Model B | Generative Approach [ Transformer ] |

*b) Model Testing:* The retrieval approach model was evaluated using the distributed testing dataset. During testing, the preprocessed questions were inputted into the trained multilingual BERT or DistilBERT model. The predicted response was compared to the designated category associated with the input question. In the context of the generative approach, the model's ability to work with new data and avoid overfitting to training data is checked. This involves providing input, i.e., question and answer that generates output or answer using the trained transformer model and calculating performance metric scores.

### E. Performance Analysis

The performance of the implemented models was analyzed using different performance metrics. Two different approaches were implemented for this research study and the results of each approach employed different performance metrics. The retrieval-based approach used a multi-class classification method and was evaluated using the micro F1-score. This metric is commonly used in classification tasks with machine learning when dealing with imbalanced datasets. It measures the weighted harmonic mean of precision and recall which are the ratios of true positives to the total predicted positives and total actual positives respectively. Similarly, the generative approach was evaluated using a widely used evaluation metric called BLEU (Bilingual Evaluation Understudy). BLEU is an evaluation metric used to measure the quality of machine translation output and it was proposed by [16] in their paper "BLEU: A Method for Automatic Evaluation of Machine Translation".

### F. Dialouge Management

Dialouge management is a critical aspect in chatbot or question answering bot to make bot effective. In this study, a rule-based approach was implemented for dialogue management. Specifically, two different rules were implemented for dialogue management. The first rule involves classifying the greetings of the user's input and responding with a specific greeting. For instance, if the user inputs "नमस्ते", the chatbot responds with "नमस्ते, म हजुरलाई कसरी सहयोग गर्न सक्छु?" . The second rule deals with handling input that is out of the chatbot's scope or intent not recognized. To accomplish this, a vocabulary or list of words was created from the training dataset and based on that the tokenized user input is checked to see if any of the words in the user's input are present in the vocabulary. If the user's input does not match any words in the vocabulary, the chatbot responds with a predefined message such as "माफ गर्नुहोला, मैले तपाईको प्रश्न बुझ्न सकिन" . This rule-based dialogue management approach enables the chatbot to respond to user input in a structured and appropriate manner, even when the input is outside the chatbot's understanding. The figure 3.15 depicts the flow of question answer bot.

## IV. Experimentation and Results

### A. Experiments

Altogether three different experiments have been performed during this study among them two was on retrieval based approach and one for generative based approach on different paramater values as shown in Table II. We initiated Experiment 1 by employing a non-stemmed dataset, fine-tuning both Multilingual BERT and DistilBERT models with specific hyperparameters, including a learning rate of 0.00003, a batch size of 32, and 20 training epochs. The results showcased remarkable performance, with Multilingual DistilBERT slightly outperforming Multilingual BERT, achieving a testing accuracy of 91.37% and a micro F1-score of 0.9165. In Experiment 2, we replicated these models using a stemmed dataset, revealing the detrimental impact of stemming on model performance. For instance, Multilingual BERT exhibited a lower testing accuracy of 75.76% and a micro F1-score of 0.8062, while Multilingual DistilBERT achieved a testing accuracy of 69.12% and a micro F1-score of 0.7362 on the stemmed dataset. Transitioning to Experiment 3, our generative approach, we employed a transformer model for a pregnancy-related question-answering bot, examining the influence of stemming. Remarkably, the transformer model excelled on the non-stemmed dataset, displaying superior language comprehension, with a training accuracy of 52.85%, validation accuracy of 52.11%, and higher BLEU scores. These results offer a nuanced understanding of model performance under different conditions, highlighted by numerical values. Please refer to Table III and Table IV for detailed performance metrics, offering insights into the effectiveness of our models in the context of chatbot development.

Table II
Model parameters used in each experiment

| Experiment | Model | Stemming | Batch Size | Learning rate | Epoch | Others |
|---|---|---|---|---|---|---|
| Exp. 1 | A | No | 32 | 0.00003 | 20 | 0.01 (decay) |
| Exp. 2 | A | Yes | 32 | 0.00003 | 20 | 0.01 (decay) |
| Exp. 3 | B | Both | 32 | 0.001 | 20 | 0.1 (dropout) |

Table III
Performance of Model A

| Model A Imp. | Stemming | Train. Accuracy | Val. Accuracy | Test Accuracy | Micro F1-Score |
|---|---|---|---|---|---|
| Multilingual BERT | Yes | 0.9842 | 0.9197 | 0.7576 | 0.8062 |
| Multilingual DistilBERT | Yes | 0.9879 | 0.9015 | 0.6912 | 0.7362 |
| Multilingual BERT | No | 0.9941 | 0.9455 | 0.9020 | 0.9157 |
| Multilingual DistilBERT | No | 0.9959 | 0.9512 | 0.9137 | **0.9165** |

Table IV
Performance of Model B on prepared dataset with and without stemming

| Model B Imp. | Stemming | Train. Loss | Val. Loss | Train. Accuracy | Val. Accuracy | BLEU-1 | BLEU-2 |
|---|---|---|---|---|---|---|---|
| Transformer | No | 0.1125 | 0.1692 | 0.5285 | 0.5211 | 0.2017 | 0.0709 |
| Transformer | Yes | 0.1360 | 0.2477 | 0.6697 | 0.6663 | 0.3570 | 0.1413 |

### *B. Findings*

The findings from performed experiments are listed below:

- In Model A architecture implementation, model with non stemmed dataset have performed well.
- The intent classification model using Multilingual BERT and DistilBERT models without a stemmed dataset demonstrated high accuracy during training, validation, and testing. This is because the training and validation loss curves during training did not exhibit a significant gap and increased in parallel indicating effective learning by the models.
- During training the loss curves of both Multilingual BERT and DistilBERT models with a stemmed dataset exhibit a notable difference between the validation loss and training loss, which indicates the occurrence of overfitting. This overfitting have led to a considerable drop in testing accuracy despite having achieved good training and validation accuracy.
- In Model B architecture implementation, model with stemmed dataset have performed well.
- The implementation of Model B architecture without a stemmed dataset has not been effective in learning resulting in a low BLEU score.
- The transformer-based model for the pregnancy-related question answering bot with stemmed dataset has achieved better performance than the previous experiment on non-stemmed dataset with similar hyperparameters, as it has better training and validation accuracy, BLEU score and lower training and validation loss. However, it's still not generating coherent and accurate answers.

### *C. Sample Output*

The sample of implemented pregnancy related domain specific chatbot with best performing DistilBERT model shows in figure 4. Similarly, the sample chat output of the generative approach in figure 5 illustrates a screenshot that serves as a representation of a model developed for a pregnancy-related chatbot within the scope of this research.

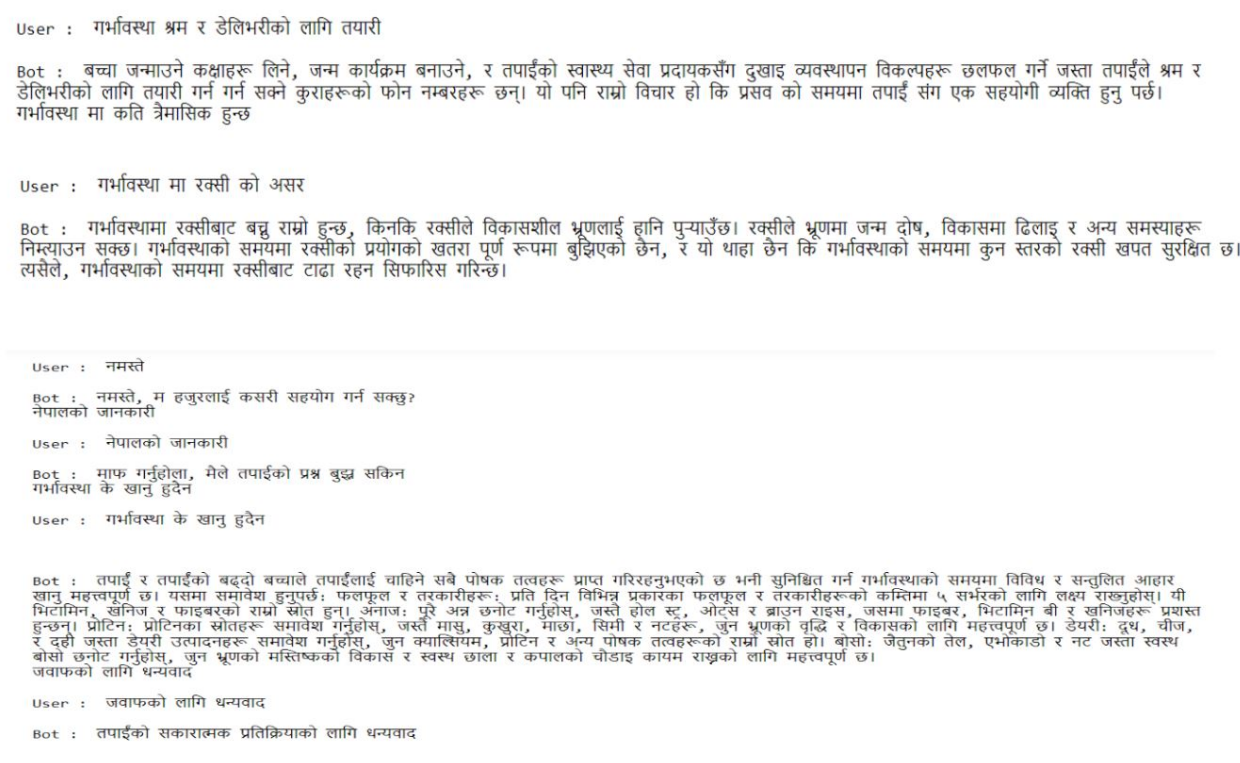

Figure 4. Sample of implemented pregnancy related domain specific chatbot with best performing DistilBERT model

```
def predict(sentence):
    prediction = evaluate(sentence)
    predicted_sentence = tokenizer.decode(
        [i for i in prediction if i < tokenizer.vocab_size]
    )
    return predicted_sentence
```

Let's test our model!

```
In [54]: predict("गर्भावस्था कति लामो हुन्छ")

Out[54]: 'गर्भावस्था सामान्यतया अन्तिम महिनावारीको पहिलो दिनदेखि लगभग ४० हप्ता वा ९ महिनासम्म रहन्छ। गर्भावस्थालाई तीन त्रैमासिकमा विभाजन गरिएको छ: पहिलो
त्रैमासिक हप्ता १ देखि १२ हप्ता सम्म, दोस्रो त्रैमासिक हप्ता १३ देखि हप्ता २६ सम्म, र तेस्रो त्रैमासिक अवधिमा २७ देखि हप्ता ४० सम्म हो।'
```

Figure 5. Sample of response from best performing generative model

## V. Discussion

The performance of transformer-based pre-trained models including Multilingual BERT and Multilingual DistilBERT through four experiments using stemmed and non-stemmed datasets for a retrieval-based pregnancy question answer bot have been evaluated. Among these, Multilingual DistilBERT without stemming displayed the best performance across all metrics, indicating its superiority. Both Multilingual BERT models without stemming also exhibited comparable or slightly lower accuracy. Interestingly, the models with stemmed datasets lagged behind those without stemming. This suggests that the BERT models' inherent ability to comprehend contextual meaning obviates the need for stemming. Consequently, the models perform well on non-stemmed data due to their ability to extract meaning from context.
Similarly, the transformer-based architectures for generative pregnancy-related question answering on stemmed and non-stemmed datasets was experimented that yields distinct outcomes. While the models demonstrated satisfactory accuracy during training and validation, the BLEU scores indicated suboptimal performance in generating coherent answers. In short, although transformer-based models exhibited promise in terms of training and validation accuracy, they struggled to generate accurate and coherent responses. Future research avenues could explore methods to enhance the models' ability to generate accurate and coherent answers.

## VI. Conclusion and Future Works

In this research, our primary objective was to develop a dataset for a pregnancy-related chatbot utilizing natural language processing. We successfully created and fine-tuned the dataset using Multilingual BERT and Multilingual DistilBERT models for retrieval-based and generative-based tasks. Our findings highlighted that Multilingual BERT and Multilingual DistilBERT performed better without stemming for retrieval-based tasks, while scratch transformer models excelled with stemmed data for generative tasks. These insights underscore the significance of effective data preprocessing in enhancing model performance.

While our study provides valuable insights, limitations exist. The dataset's comprehensiveness could be improved, and alternative data augmentation techniques, like back-translation, might enhance diversity. Addressing translation challenges through advanced methods and refining data quality are areas for future exploration. Additionally, training models with pre-trained word embedding weights and integrating the chatbot across platforms could further elevate its utility and impact..